\def\year{2019}\relax
\documentclass[letterpaper]{article} %DO NOT CHANGE THIS
\usepackage{aaai19}  %Required
\usepackage{times}  %Required
\usepackage{helvet}  %Required
\usepackage{courier}  %Required
\usepackage{url}  %Required
\usepackage{graphicx}  %Required

\usepackage{amssymb} 
\usepackage{pifont}
\usepackage{amsmath}
\usepackage{bm}
\usepackage{multirow}
\usepackage{subcaption}

\newcommand{\cmark}{\ding{51}}%
\newcommand{\xmark}{\ding{55}}%
\newcommand{\vect}[1]{\bm{#1}}
\newcommand{\mat}[1]{\bf{#1}}
\newcommand{\RR}{\mathbb{R}}

\begin{document}
\title{Sequential Attention-based Network for Noetic End-to-End Response Selection}
\author{Qian Chen, Wen Wang \\
Speech Lab, DAMO Academy, Alibaba Group \\
\{tanqing.cq,~w.wang\}@alibaba-inc.com
}
\maketitle
\begin{abstract}
The noetic end-to-end response selection challenge as one track in Dialog System Technology Challenges 7 (DSTC7) aims to push the state of the art of utterance classification for real world goal-oriented dialog systems, for which participants need to select the correct next utterances from a set of candidates for the multi-turn context. This paper describes our systems that are ranked the top on both datasets under this challenge, one focused and small (Advising) and the other more diverse and large (Ubuntu). Previous state-of-the-art models use hierarchy-based (utterance-level and token-level) neural networks to explicitly model the interactions among different turns' utterances for context modeling. In this paper, we investigate a sequential matching model based only on chain sequence for multi-turn response selection. Our results demonstrate that the potentials of sequential matching approaches have not yet been fully exploited in the past for multi-turn response selection.  In addition to ranking the top in the challenge, the proposed model outperforms all previous models, including state-of-the-art hierarchy-based models, and achieves new state-of-the-art performances on two large-scale public multi-turn response selection benchmark datasets.

\end{abstract}
\section{Introduction}
Dialogue systems are gaining more and more attention due to their encouraging potentials and commercial values. With the recent success of deep learning models~\cite{DBLP:conf/aaai/SerbanSBCP16}, building an end-to-end dialogue system became feasible. However, building end-to-end multi-turn dialogue systems is still quite challenging, requiring the %system to remember and comprehend multi-turn 
system to memorize and comprehend multi-turn
conversation context, rather than only considering the current utterance as in single-turn dialogue systems.

Multi-turn dialogue modeling can be divided into generation-based methods~\cite{DBLP:conf/aaai/SerbanSBCP16,DBLP:conf/aaai/ZhouLCLCH17} and retrieval-based methods~\cite{DBLP:conf/sigdial/LowePSP15,DBLP:conf/acl/WuWXZL17}. The latter is the focus of the noetic end-to-end response selection challenge in DSTC7\footnote{http://workshop.colips.org/dstc7/}~\cite{DSTC7}. Retrieval-based methods select the best response from a candidate pool for the multi-turn context, which can be considered as performing a multi-turn response selection task.
The typical approaches for multi-turn response selection mainly consist of sequence-based methods~\cite{DBLP:conf/sigdial/LowePSP15,DBLP:conf/sigir/YanSW16} and hierarchy-based methods~\cite{DBLP:conf/emnlp/ZhouDWZYTLY16,DBLP:conf/acl/WuWXZL17,DBLP:conf/coling/ZhangLZZL18,DBLP:conf/acl/WuLCZDYZL18}. Sequence-based methods usually concatenate the context utterances into a long sequence. Hierarchy-based methods normally model each utterance individually and then explicitly model the interactions among the utterances. 

Recently, previous work \cite{DBLP:conf/acl/WuWXZL17,DBLP:conf/coling/ZhangLZZL18} claims that hierarchy-based methods with complicated networks can achieve significant gains over sequence-based methods. However, in this paper, we investigate the efficacy of a sequence-based method, i.e., Enhanced Sequential Inference Model (ESIM)~\cite{DBLP:conf/acl/ChenZLWJI17} originally developed for the natural language inference (NLI) task. Our systems are ranked the top on both datasets, i.e., Advising and Ubuntu datasets, under the DSTC7 response selection challenge. In addition, the proposed approach outperforms all previous models, including the previous state-of-the-art hierarchy-based methods, on two large-scale public benchmark datasets, the Lowe's Ubuntu~\cite{DBLP:conf/sigdial/LowePSP15} and
E-commerce datasets~\cite{DBLP:conf/coling/ZhangLZZL18}. Our source code is available at \url{https://github.com/alibaba/esim-response-selection}.

Hierarchy-based methods usually use extra neural
networks to explicitly model the multi-turn utterances' relationship. They also usually need to truncate the utterances in the multi-turn context to make them the same length and shorter than the maximum length. However, the lengths of different turns usually vary significantly in real tasks. When using a large maximum length, we need to add a lot of zero padding in hierarchy-based methods, which will increase computation complexity and memory cost drastically. When using a small maximum length, we may throw away some important information in the multi-turn context. We propose to use a sequence-based model, the ESIM model, in the multi-turn response selection task to effectively address the above mentioned problem encountered by hierarchy-based methods. We concatenate the multi-turn context as a long sequence, and convert the multi-turn response selection task into a sentence pair binary classification task, i.e., whether the next sentence is the response for the current context. There are two major advantages of ESIM over hierarchy-based methods. First, since ESIM does not need to make each utterance the same length, it has less zero padding and hence could be more computationally efficient than hierarchy-based methods. Second, ESIM models the interactions between utterances in the context implicitly, yet in an effective way as described in the model description section, without 
using extra complicated networks.

\section{Task Description}

DSTC7 is divided into 3 different tracks, and the proposed approach is developed for the noetic end-to-end response selection track. This track focuses on goal-oriented multi-turn dialogs and the objective is to select the correct response from a set of candidates. Participating systems should not be based on hand-crafted features or rule-based systems. Two datasets are provided, i.e., Ubuntu and Advising, which will be introduced in detail in the experiment section.

The response selection track provided series of subtasks that have similar structures, but vary in the output space and available context. In Table~\ref{tab:stat:task}, \cmark~indicates that the task is evaluated on the marked dataset, and \xmark~indicates not applicable. 

\begin{table*}[ht]
\begin{center}
\scalebox{0.9}{
\begin{tabular}{c l c c}
\hline
\multicolumn{1}{l}{\textbf{Sub}} & \multicolumn{1}{l}{\textbf{Description}} & \multicolumn{1}{l}{\textbf{Ubuntu}}  & \multicolumn{1}{l}{\textbf{Advising}}  \\
\hline
1 &Select the next utterance from a candidate pool of 100 sentences	  &\cmark & \cmark\\
2 &Select the next utterance from a candidate pool of 120000 sentences &\cmark & \xmark\\
3 &Select the next utterance and its paraphrases from a candidate pool of 100 sentences &\xmark & \cmark \\
4 &Select the next utterance from a candidate pool of 100 which might not contain the correct next utterance & \cmark & \cmark\\
5 & Select the next utterance from a candidate pool of 100 incorporating the external knowledge & \cmark & \cmark\\
\hline
\end{tabular}
}
\end{center}
\caption{Task description.}
\label{tab:stat:task}
\end{table*}

\section{Model Description}

The multi-turn response selection task is to select the next utterance from a candidate pool, given a multi-turn context. We convert the problem into a binary classification task, similar to the previous work~\cite{DBLP:conf/sigdial/LowePSP15,DBLP:conf/acl/WuWXZL17}. Given a multi-turn context and a candidate response, our model needs to determine whether or not the candidate response is the correct next utterance. In this section, we will introduce our model, Enhanced Sequential Inference Model (ESIM) \cite{DBLP:conf/acl/ChenZLWJI17} originally developed for natural language inference. The model consists of three main components, i.e., input encoding, local matching, and matching composition, as shown in Figure 1(b).

\begin{figure*}[!t]
\centering
\begin{subfigure}[b]{0.49\textwidth}
\centering
\includegraphics[width=\textwidth]{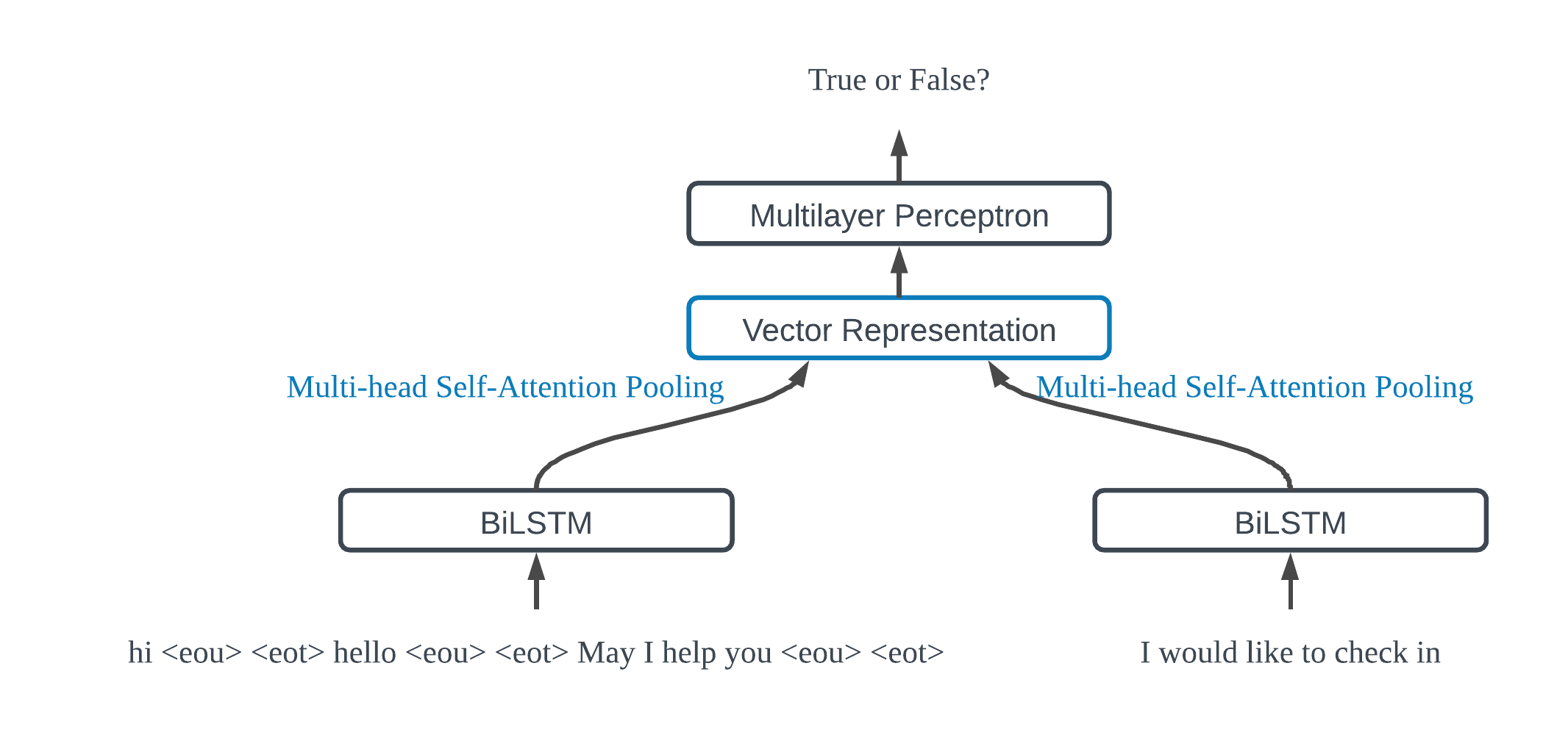}
\caption{Sentence encoding-based method.}
\label{fig:sent_enc}
\end{subfigure}
~ %add desired spacing between images, e. g. ~, \quad, \qquad, \hfill etc. 
\begin{subfigure}[b]{0.49\textwidth}
\centering
\includegraphics[width=\textwidth]{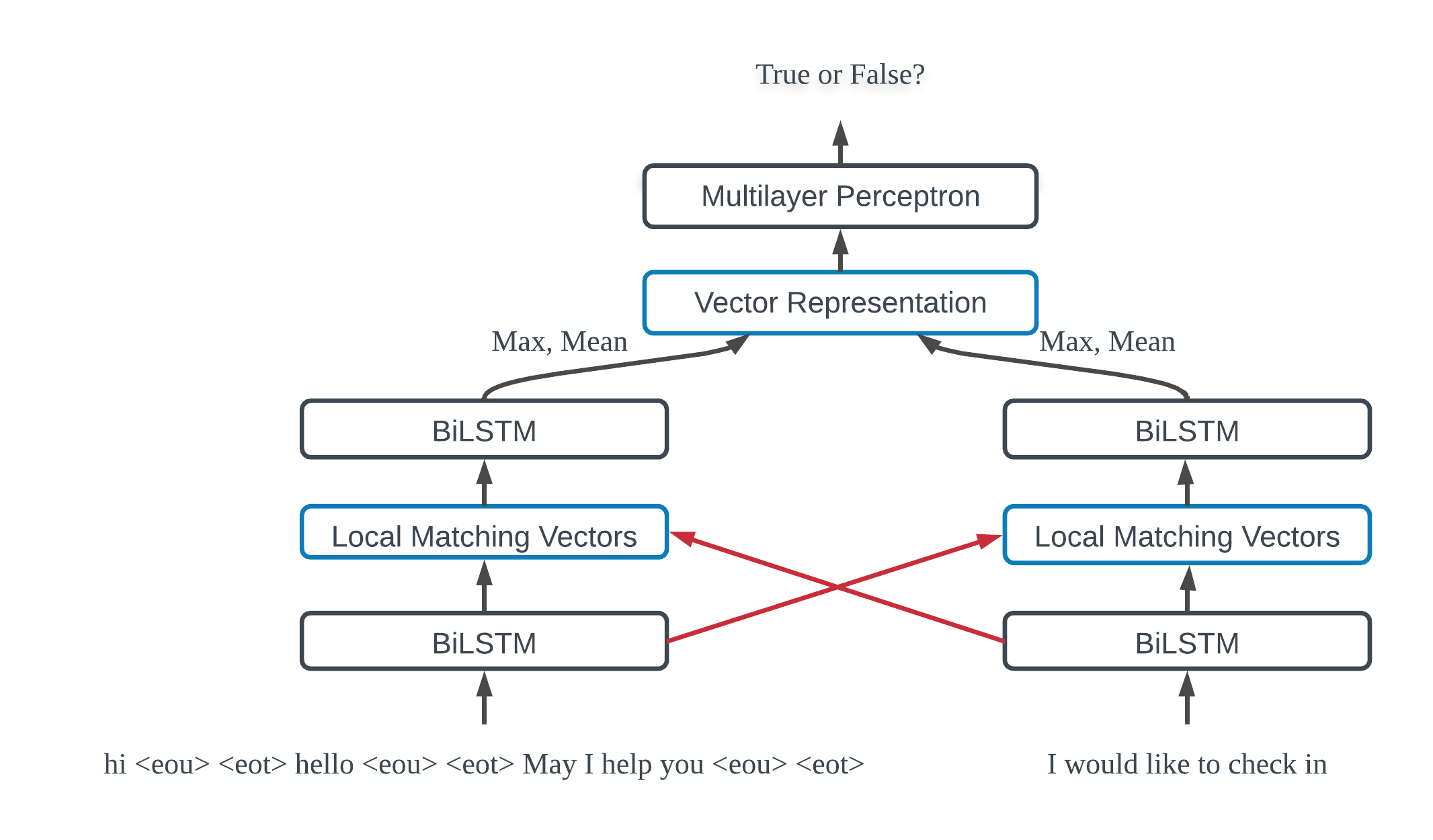}
\caption{Cross attention-based method.}
\label{fig:cross_att}	
\end{subfigure}
\caption{Two kinds of neural network-based methods for sentence pair classification.}
\label{fig_model}
\end{figure*}

\subsection{Input Encoding}
\label{sec:input}
Input encoding encodes the context information and represents tokens in their contextual meanings. Instead of encoding the context information through complicated hierarchical structures as in hierarchy-based methods, ESIM encodes the context information simply as follows. The multi-turn context is concatenated as a long sequence, which is denoted as $\vect{c} = (c_1,\dots,c_m)$. The candidate response is denoted as $\vect{r} = (r_1,\dots,r_{n})$. Pre-trained word embedding ${\mat E} \in \RR^{d_e \times |V|}$ is then used to convert $\vect{c}$ and $\vect{r}$ to two vector sequences $[{\mat E}(c_1),\dots,{\mat E}(c_m)]$ and $[{\mat E}(r_1),\dots,{\mat E}(r_n)]$, where $|V|$ is the vocabulary size and $d_e$ is the dimension of the word embedding. There are many kinds of pre-trained word embeddings available, such as GloVe \cite{DBLP:conf/emnlp/PenningtonSM14} and fastText \cite{DBLP:conf/lrec/MikolovGBPJ18}. We propose a method to exploit multiple embeddings. Given $k$ kinds of pre-trained word embeddings ${\mat E}_1, \dots, {\mat E}_k$, we concatenate all embeddings for the word $i$, i.e., ${\mat E}(c_i) = [{\mat E}_1(c_i);\dots;{\mat E}_k(c_i)]$. Then we use a feed-forward layer with ReLU to reduce dimension from $(d_{e_1}+\dots+d_{e_k})$ to $d_h$.

To represent tokens in their contextual meanings, the context and the response are fed into BiLSTM encoders to obtain context-dependent hidden states ${\vect c}^s$ and ${\vect r}^s$:
\begin{align}
{\vect c}^s_i &=\mathrm{BiLSTM}_1({\mat E}(\vect{c}),i) \,,\\
{\vect r}^s_j &=\mathrm{BiLSTM}_1({\mat E}(\vect{r}),j) \,,
\end{align}
\noindent where $i$ and $j$ indicate the $i$-th token in the context and the $j$-th token in the response, respectively.

\subsection{Local Matching}
Modeling the local semantic relation between a context and a response is the critical component for determining whether the response is the proper next utterance. For instance, a proper response usually relates to some keywords in the context, which can be obtained by modeling the local semantic relation. 
Instead of directly encoding the context and the response as two dense vectors, we use the cross-attention mechanism to align the tokens from the context and response, and then calculate the semantic relation at the token level. The attention weight is calculated as:
\begin{equation}
e_{ij} = ({\vect c}^s_i)^\mathrm{T} {\vect r}^s_j \,.
\label{eq:eij}
\end{equation}

Soft alignment is used to obtain the local relevance between the context and the response, which is calculated by the attention matrix ${\mat e} \in \RR^{m \times n}$  in Equation~(\ref{eq:eij}). Then for the hidden state of the $i$-th token in the context, i.e., ${\vect c}^s_i$ (already encoding the token itself and its contextual meaning), the relevant semantics in the candidate response is 
identified as a vector ${\vect c}^d_i$, called dual vector here, which is a weighted combination of all the response's states, more specifically as shown in Equation~(\ref{eq:a_c}).
\begin{align}
\label{eq:a_c}
\alpha_{ij} & = \frac{\exp(e_{ij})}{\sum_{k=1}^{n}\exp(e_{ik})} \,, ~{\vect c}^d_i =\sum_{j=1}^{n}\alpha_{ij} {\vect r}^s_j \,,\\
\label{eq:b_c}
\beta_{ij} & = \frac{\exp(e_{ij})}{\sum_{k=1}^{m}\exp(e_{kj})} \,, ~{\vect r}^d_j =\sum_{i=1}^{m}\beta_{ij} {\vect c}^s_i \,,
\end{align}
\noindent where ${\vect \alpha} \in \RR^{m \times n}$ and ${\vect \beta} \in \RR^{m \times n}$ are the normalized attention weight matrices with respect to the $2$-axis and $1$-axis. The similar calculation is performed for the hidden state of each token in the response, i.e., ${\vect r}^s_j$, 
%with
as in
Equation~(\ref{eq:b_c}) to obtain the dual vector ${\vect r}^d_j$. 

By comparing vector pair $<{\vect c}_i^s, {\vect c}_i^d>$, we can model the token-level semantic relation between aligned token pairs. The similar calculation is also applied for vector pair $<{\vect r}_j^s,{\vect r}_j^d>$. We collect local matching information as follows:
% {\fontsize{10pt}{1.0cm}
\begin{align}
\label{eq:infer1}
{\vect c}^l_i &= F([{\vect c}^s_i;{\vect c}^d_i;{\vect c}^s_i - {\vect c}^d_i ;{\vect c}^s_i \odot {\vect c}^d_i]) \,,\\
\label{eq:infer2}
{\vect r}^l_j &= F([{\vect r}^s_j;{\vect r}^d_j;{\vect r}^s_j - {\vect r}^d_j; {\vect r}^s_j \odot {\vect r}^d_j]) \,,
\end{align}
\noindent where a heuristic matching approach~\cite{DBLP:conf/acl/MouMLX0YJ16} with difference and element-wise product is used here to obtain local matching vectors ${\vect c}^l_i$ and ${\vect r}^l_j$ for the context and response, respectively. 
$F$ is a one-layer feed-forward neural network with ReLU to reduce the dimension. 

\subsection{Matching Composition}
Matching composition is realized as follows. To determine whether the response is the next utterance for the current context, we explore a composition layer to compose the local matching vectors (${\vect c}^l$ and ${\vect r}^l$) collected above:
\begin{align}
{\vect c}^v_i &= \mathrm{BiLSTM}_2({\vect c}^l,i) \,, \\
{\vect r}^v_j &= \mathrm{BiLSTM}_2({\vect r}^l,j) \,.
\end{align}
Again we use BiLSTMs as building blocks for the composition layer, but the role of BiLSTMs here is completely different from that in the input encoding layer. The BiLSTMs here read local matching vectors (${\vect c}^l$ and ${\vect r}^l$) and learn to discriminate critical local matching vectors for the overall utterance-level relationship. 

The output hidden vectors of $\mathrm{BiLSTM}_2$ are converted to fixed-length vectors through pooling operations and fed to the final classifier to determine the overall relationship. Max and mean poolings are used and concatenated altogether to obtain a fixed-length vector. Then the final vector is fed to the multi-layer perceptron (MLP) classifier, with one hidden layer, \textit{tanh} activation, and \textit{softmax} output layer. The entire ESIM model is trained via minimizing the cross-entropy loss in an end-to-end manner. 
\begin{align}
y &= \mathrm{MLP}([{\vect c}^v_{max};{\vect c}^v_{mean};{\vect r}^v_{max};{\vect r}^v_{mean}]) \,.
\end{align}

\subsection{Sentence-encoding based methods}
For subtask 2 on the Ubuntu dataset, we need to select the next utterance from a candidate pool of 120000 sentences. If we use the cross-attention based ESIM model directly, the computation cost is unacceptable. Instead, we first use a sentence-encoding based method to select the top 100 candidates from 120000 sentences and then rerank them using ESIM. Sentence-encoding based models use the Siamese architecture~\cite{DBLP:conf/nips/BromleyGLSS93,DBLP:conf/repeval/ChenZLWJI17} shown in Figure 1 (a).  Parameter-tied neural networks are applied to encode both the context and the response. Then a neural network classifier is applied to decide the relationship between the two sentences. Here, we use BiLSTMs with multi-head self-attention pooling to encode sentences~\cite{DBLP:journals/corr/LinFSYXZB17,DBLP:conf/coling/ChenLZ18}, and an MLP to classify.

We use the same input encoding process as ESIM. To transform a variable length sentence into a fixed length vector representation, we use a weighted summation of all BiLSTM hidden vectors (${\mat H}$): 
\begin{align}
{\mat A} = \text{softmax}({\mat W}_2\text{ReLU}({\mat W}_1 {\mat H}^\mathrm{T} + {\vect b_1} ) + {\vect b_2})^\mathrm{T} \,,
\end{align} 
\noindent where ${\mat W}_1 \in \RR^{d_a \times 2d_h}$ and ${\mat W}_2 \in \RR^{d_m \times d_a}$ are weight matrices; ${\vect b_1} \in \RR^{d_a}$ and ${\vect b_2} \in \RR^{d_m}$ are bias; $d_a$ is the dimension of the attention network and $d_h$ is the dimension of BiLSTMs. ${\mat H} \in \RR^{T \times 2d_h}$ are the hidden vectors of BiLSTMs, where $T$ denotes the length of the sequence. ${\mat A} \in \RR^{T \times d_m}$ is the multi-head attention weight matrix, where $d_m$ is a hyperparameter of the head number that needs to be tuned using the held-out set. Instead of using max pooling or mean pooling, we sum up the BiLSTM hidden states ${\mat H}$ according to the weight matrix ${\mat A}$ to get a vector representation of the input sentence.
\begin{align}
{\mat V} =  {\mat A}^\mathrm{T}{\mat H} \,,
\end{align} 
\noindent where the matrix ${\mat V} \in \RR^{d_m \times 2d_h}$ can be flattened into a vector representation ${\vect v} \in \RR^{2d_h d_m }$. To enhance the relationship between sentence pairs, similarly to ESIM, we concatenate the embeddings of two sentences and their absolute difference and element-wise product~\cite{DBLP:conf/acl/MouMLX0YJ16} as the input to the MLP classifier:
\begin{equation}
y = \mathrm{MLP}([{\vect v}_c;{\vect v}_r;\lvert{\vect v}_c - {\vect v}_r\rvert; {\vect v}_c \odot {\vect v}_r]) \,.
\end{equation}
 
The MLP has two hidden layers with \textit{ReLU} activation, shortcut connections, and \textit{softmax} output layer. The entire model is trained end-to-end through minimizing the cross-entropy loss. 

\section{Experiments}
\subsection{Datasets}
We evaluated our model on both datasets of the DSTC7 response selection track, i.e., the Ubuntu and Advising datasets. In addition, to compare with previous methods, we also evaluated our model on two large-scale public muli-turn response selection benchmarks, i.e., the Lowe's Ubuntu dataset~\cite{DBLP:conf/sigdial/LowePSP15} and E-commerce dataset~\cite{DBLP:conf/coling/ZhangLZZL18}.
% Data statistics are summarized in Table~\ref{tab:stat}.
\subsubsection{Ubuntu dataset}
The Ubuntu dataset includes two party conversations from Ubuntu Internet Relay Chat (IRC) channel~\cite{kummerfeld2018analyzing}. Under this challenge, the context of each dialog contains more than 3 turns and the system is asked to select the next turn from the given set of candidate sentences. Linux manual pages are also provided as external knowledge. We used a similar data augmentation strategy as in \cite{DBLP:conf/sigdial/LowePSP15}, i.e., we considered each utterance (starting at the second one) as a potential response, with the previous utterances as its context. Hence a dialogue of length 10 yields 9 training examples. To train a binary classifier, we need to sample negative responses from the candidate pool. Initially, we used a 1:1 ratio between positive and negative responses for balancing the samples. Later, we found using more negative responses improved the results, such as 1:4 or 1:9. Considering efficiency, we chose 1:4 in the final configuration for all subtasks except 1:1 for subtask 2.

\subsubsection{Advising dataset}
The advising dataset includes two-party dialogs that simulate a discussion between a student and an academic advisor. Structured information is provided as a database including course information and personas. The data also includes paraphrases of the sentences and the target responses. We used a similar data augmentation strategy as for the Ubuntu dataset based on original dialogs and their paraphrases. The ratio between positive and negative responses is 1:4.33.

\subsubsection{Lowe's Ubuntu dataset}
This dataset is similar to the DSTC7 Ubuntu data. The training set contains one million context-response pairs and the ratio between positive and negative responses is 1:1. On both development and test sets, each context is associated with one positive response and 9 negative responses. 

\subsubsection{E-commerce dataset}
The E-commerce dataset~\cite{DBLP:conf/coling/ZhangLZZL18} is collected from real-word conversations between customers and customer service staff from Taobao\footnote{https://www.taobao.com}, the largest e-commerce platform in China. The ratio between positive and negative responses is 1:1 in both training and development sets, and 1:9 in the test set. 

\subsection{Training Details}
We used spaCy\footnote{https://spacy.io/} to tokenize text for the two DSTC7 datasets, and used original tokenized text without any further pre-processing for the two public datasets. The multi-turn context was concatenated and two special tokens, \_\_eou\_\_ and \_\_eot\_\_, were inserted, where \_\_eou\_\_ denotes end-of-utterance and \_\_eot\_\_ denotes end-of-turn. 

The hyperparameters were tuned based on the development set.
We used GloVe~\cite{DBLP:conf/emnlp/PenningtonSM14} and fastText~\cite{DBLP:conf/lrec/MikolovGBPJ18} as pre-trained word embeddings. For subtask 5 of the Ubuntu dataset, we also used word2vec~\cite{DBLP:conf/nips/MikolovSCCD13} to train word embedding from the provided Linux manual pages. The detail is shown in Table \ref{tab:stat:word}. 
Note that for subtask 5 of the Advising dataset, we tried using the suggested course information as external knowledge but didn't observe any improvement. Hence, we submitted the results for the Advising dataset without using any external knowledge. 
For Lowe's Ubuntu and E-commerce datasets, we used pre-trained word embedding on the training data by word2vec~\cite{DBLP:conf/nips/MikolovSCCD13}. 
The pre-trained embeddings were fixed during the training procedure for the two DSTC7 datasets, but fine-tuned for Lowe's Ubuntu and E-commerce datasets. 

Adam~\cite{DBLP:journals/corr/KingmaB14} was used for optimization with an initial learning rate of $0.0002$ for Lowe's Ubuntu dataset, and $0.0004$ for the rest. The mini-batch size was set to $128$ for DSTC7 datasets, $16$ for the Lowe's Ubuntu dataset, and $32$ for the E-commerce dataset. The hidden size of BiLSTMs and MLP was set to 300.

To make the sequences shorter than the maximum length, we cut off last tokens for the response but did the cut-off in the reverse direction for the context, as we hypothesized that the last few utterances in the context is more important than the first few utterances.
For the Lowe's Ubuntu dataset, the maximum lengths of the context and response were set to 400 and 150, respectively; for the E-commerce dataset, 300 and 50; for the rest datasets, 300 and 30. 

More specially, for subtask 2 of DSTC7 Ubuntu, we used BiLSTM hidden size 400 and 4 heads for sentence-encoding methods. For subtask 4, the candidate pool may not contain the correct next utterance, so we need to choose a threshold. When the probability of positive labels is smaller than the threshold, we predict that candidate pool doesn't contain the correct next utterance. The threshold was selected from the range $[0.50,0.51,..,0.99]$ based on the development set. 

\begin{table}[ht]
\begin{center}
\scalebox{0.9}{
\begin{tabular}{l l r}
\hline
\multicolumn{1}{l}{\textbf{Embedding}} & \multicolumn{1}{l}{\textbf{Training corpus}} & \multicolumn{1}{l}{\textbf{\#Words}}   \\
\hline
glove.6B.300d & Wikipedia + Gigaword  & 0.4M \\
glove.840B.300d & Common Crawl   & 2.2M\\
glove.twitter.27B.200d & Twitter &  1.2M\\
wiki-news-300d-1M.vec & Wikipedia + UMBC  & 1.0M \\
crawl-300d-2M.vec & Common Crawl  &  2.0M \\
word2vec.300d & Linux manual pages & 0.3M \\
\hline
\end{tabular}
}
\end{center}
\caption{Statistics of pre-trained word embeddings. The 1-3 rows are from GloVe; 4-5 rows are from fastText; 6 is from word2vec.}
\label{tab:stat:word}
\end{table}

\subsection{Results}
Our results on all DSTC7 response selection subtasks were summarized in Table~\ref{tab:result}. The challenge ranking considers the average of Recall@10 and Mean Reciprocal Rank (MRR). On the Advising dataset, the test case 2 (Advising2) results were considered for ranking, because test case 1 (Advising1) has some dependency on the training dataset. Our results rank first on 7 subtasks, rank second on subtask 2 of Ubuntu, and overall rank first on both datasets of the DSTC7 response selection challenge\footnote{The official evaluation allows up to 3 different settings, but we only submitted one setting.}. Subtask 3 may contain multiple correct responses, so Mean Average Precision (MAP) is considered as an extra metric.

\begin{table}[t]
\begin{center}
\scalebox{0.9}{
\begin{tabular}{l l r r r}
\hline
\multicolumn{1}{l}{\textbf{Subtask}} & \multicolumn{1}{l}{\textbf{Measure}} & \multicolumn{1}{l}{\textbf{Ubuntu}} & \multicolumn{1}{l}{\textbf{Advising1}} & \multicolumn{1}{l}{\textbf{Advising2}}  \\
\hline
\multirow{4}{*}{Subtask1} & Recall@1 & 0.645 & 0.398 & 0.214 \\
& Recall@10 & 0.902 & 0.844 & 0.630 \\
& Recall@50 & 0.994 & 0.986 & 0.948 \\
& MRR & 0.735 & 0.5408 & 0.339 \\
\hline
\multirow{4}{*}{Subtask2} & Recall@1 & 0.067 & \multicolumn{2}{c}{\multirow{4}{*}{NA}} \\
& Recall@10 & 0.185 & & \\
& Recall@50 & 0.266 & & \\
& MRR & 0.1056 & & \\
\hline
\multirow{5}{*}{Subtask3}& Recall@1 & \multirow{5}{*}{NA}  & 0.476 & 0.290 \\
& Recall@10 & & 0.906 & 0.750 \\
& Recall@50 & & 0.996 & 0.978 \\
& MRR & & 0.6238 & 0.4341 \\
& MAP & & 0.7794 & 0.5327 \\
\hline
\multirow{4}{*}{Subtask4} & Recall@1 & 0.624 & 0.372 & 0.232 \\
& Recall@10 & 0.941 & 0.886 & 0.692 \\
& Recall@50 & 0.997 & 0.990 & 0.938 \\
& MRR & 0.742 & 0.5409 & 0.3826 \\
\hline
\multirow{4}{*}{Subtask5} & Recall@1 & 0.653 & 0.398 & 0.214 \\
& Recall@10 & 0.905 & 0.844 & 0.630 \\ 
& Recall@50 & 0.995 & 0.986 & 0.948 \\
& MRR & 0.7399 & 0.5408 & 0.339 \\
\hline

\end{tabular}
}
\end{center}
\caption{The submission results on the hidden test sets for the DSTC7 response selection challenge. NA - not applicable. In total, there are 8 test conditions.}
\label{tab:result}
\end{table}

\subsection{Ablation Analysis}

\begin{table}[t]
\begin{center}
\scalebox{0.9}{
\begin{tabular}{l l r r r r}
\hline
\multicolumn{1}{l}{\textbf{Sub}} & \multicolumn{1}{l}{\textbf{Models}} & \multicolumn{1}{l}{\textbf{R@1}} & \multicolumn{1}{l}{\textbf{R@10}} & \multicolumn{1}{l}{\textbf{R@50}} & \multicolumn{1}{l}{\textbf{MRR}}  \\
\hline
\multirow{4}{*}{1} & ESIM & 0.534 & 0.854 & 0.985 & 0.6401  \\
& -CtxDec & 0.508 & 0.845 & 0.982 & 0.6210  \\
& -CtxDec \& -Rev  & 0.504 & 0.840 & 0.982 & 0.6174 \\
& Ensemble  & 0.573 & 0.887 & 0.989 & 0.6790  \\
\hline
\multirow{5}{*}{2} & Sent-based & 0.021  & 0.082 & 0.159 & 0.0416   \\
& Ensemble1 & 0.023 & 0.091 & 0.168 & 0.0475  \\
& ESIM & 0.043 & 0.125 & 0.191 & 0.0713   \\
& -CtxDec  & 0.034 & 0.117 & 0.191 & 0.0620 \\
& Ensemble2  & 0.048 & 0.134 & 0.194 & 0.0770  \\
\hline
\multirow{4}{*}{4} & ESIM & 0.515 & 0.887 & 0.988 & 0.6434  \\
& -CtxDec & 0.492 & 0.877 & 0.987 & 0.6277  \\
& -CtxDec \& -Rev  & 0.490 & 0.875 & 0.986 & 0.6212 \\
& Ensemble  & 0.551 & 0.909 & 0.992 & 0.6771  \\
\hline
\multirow{4}{*}{5} & ESIM & 0.534 & 0.854 & 0.985 & 0.6401  \\
& +W2V & 0.530 & 0.858 & 0.986 & 0.6394  \\
& Ensemble  & 0.575 & 0.890 & 0.989 & 0.6817  \\
\hline
\end{tabular}
}
\end{center}
\caption{Ablation analysis on the development set for the DSTC7 Ubuntu dataset. }
\label{tab:result:ubuntu}
\end{table}

\begin{table}[ht]
\begin{center}
\scalebox{0.9}{
\begin{tabular}{l l r r r r}
\hline
\multicolumn{1}{l}{\textbf{Sub}} & \multicolumn{1}{l}{\textbf{Models}} & \multicolumn{1}{l}{\textbf{R@1}} & \multicolumn{1}{l}{\textbf{R@10}} & \multicolumn{1}{l}{\textbf{R@50}} & \multicolumn{1}{l}{\textbf{MRR}}  \\
\hline
\multirow{3}{*}{1} & -CtxDec & 0.222 & 0.656 & 0.954 & 0.3572  \\
& -CtxDec \& -Rev  & 0.214 & 0.658 & 0.942 & 0.3518 \\
& Ensemble  & 0.252 & 0.720 & 0.960 & 0.4010  \\
\hline
\multirow{3}{*}{3} & -CtxDec  & 0.320 & 0.792 & 0.978 & 0.4704 \\
& -CtxDec \& -Rev  & 0.310 & 0.788 & 0.978 & 0.4550 \\
& Ensemble  & 0.332 & 0.818 & 0.984 & 0.4848  \\
\hline
\multirow{3}{*}{4} & -CtxDec  & 0.248 & 0.706 & 0.970 & 0.3955  \\
& -CtxDec \& -Rev  & 0.226 & 0.714 & 0.946 & 0.3872  \\
& Ensemble  & 0.246 & 0.760 & 0.970 & 0.4110   \\
\hline
\end{tabular}
}
\end{center}
\caption{Ablation analysis on the development set for the DSTC7 Advising dataset. }
\label{tab:result:advising}
\end{table}

\begin{table*}[hbt!]
\begin{center}
\scalebox{0.9}{
\begin{tabular}{l | l l l| l l l}
\hline
\multicolumn{1}{l|}{\textbf{Models}} & \multicolumn{3}{c}{\textbf{Ubuntu}} &  \multicolumn{3}{|c}{\textbf{E-commerce}} \\
& R@1 & R@2  & R@5  & R@1  & R@2  & R@5  \\
\hline
TF-IDF~\cite{DBLP:conf/sigdial/LowePSP15} & 0.410 & 0.545 & 0.708 & 0.159 & 0.256 & 0.477 \\
RNN~\cite{DBLP:conf/sigdial/LowePSP15} & 0.403 & 0.547 & 0.819 & 0.325 & 0.463 & 0.775 \\
CNN~\cite{DBLP:journals/corr/KadlecSK15} & 0.549 & 0.684 & 0.896 & 0.328 & 0.515 & 0.792 \\
LSTM~\cite{DBLP:journals/corr/KadlecSK15} & 0.638 & 0.784 & 0.949 & 0.365 & 0.536 & 0.828 \\
BiLSTM~\cite{DBLP:journals/corr/KadlecSK15} & 0.630 & 0.780 & 0.944 & 0.355 & 0.525 & 0.825 \\
\hline
MV-LSTM~\cite{DBLP:conf/aaai/WanLGXPC16} & 0.653 & 0.804 & 0.946 & 0.412 & 0.591 & 0.857 \\
Match-LSTM~\cite{DBLP:conf/naacl/WangJ16} & 0.653 & 0.799 & 0.944 & 0.410 & 0.590 & 0.858 \\
Attentive-LSTM~\cite{DBLP:journals/corr/TanXZ15} & 0.633 & 0.789 & 0.943 &  0.401 & 0.581 & 0.849 \\
Multi-Channel~\cite{DBLP:conf/acl/WuWXZL17} & 0.656 & 0.809 & 0.942 & 0.422 & 0.609 & 0.871 \\
\hline
Multi-View~\cite{DBLP:conf/emnlp/ZhouDWZYTLY16} & 0.662 & 0.801 & 0.951 & 0.421 & 0.601 & 0.861 \\
DL2R~\cite{DBLP:conf/sigir/YanSW16} & 0.626 & 0.783 & 0.944 & 0.399 & 0.571 & 0.842 \\
SMN~\cite{DBLP:conf/acl/WuWXZL17} & 0.726 & 0.847 & 0.961 & 0.453 & 0.654 & 0.886 \\
DUA~\cite{DBLP:conf/coling/ZhangLZZL18} & 0.752 & 0.868 & 0.962 & 0.501 & 0.700 & 0.921 \\
DAM~\cite{DBLP:conf/acl/WuLCZDYZL18}  & 0.767 & 0.874 & 0.969  & - & - & - \\
\hline
Our ESIM & \textbf{0.796}& \textbf{0.894} & \textbf{0.975}   &  \textbf{0.570} & \textbf{0.767} & \textbf{0.948} \\
\hline
\end{tabular}
}
\end{center}
\vspace{-2mm}
\caption{Comparison of different models on two large-scale public benchmark datasets. All the results except ours are cited from previous work~\cite{DBLP:conf/coling/ZhangLZZL18,DBLP:conf/acl/WuLCZDYZL18}.} 
\label{tab:result:public}
\end{table*}

Ablation analysis is shown in Table~\ref{tab:result:ubuntu} and~\ref{tab:result:advising} for the Ubuntu and Advising datasets, respectively. For Ubuntu subtask 1, ESIM achieved 0.854 R@10 and 0.6401 MRR. If we removed context's local matching and matching composition to accelerate the training process (``-CtxDec''), R@10 and MRR dropped to 0.845 and 0.6210. Further discarding the last words instead of the preceding words for the context (``-CtxDec \& -Rev'') degraded R@10 and MRR to 0.840 and 0.6174. Ensembling the above three models (``Ensemble'') achieved 0.887 R@10 and 0.6790 MRR. Ensembling was performed by averaging output from models trained with different parameter initializations and different structures. 

For Ubuntu subtask 2, the sentence-encoding based methods (``Sent-based'') achieved 0.082 R@10 and 0.0416 MRR. After ensembling several models with different parameter initializations (``Ensemble1''), R@10 and MRR were increased to 0.091 and 0.0475. Using ESIM to rerank the top 100 candidates predicted by ``Ensemble1'' achieved 0.125 R@10 and 0.0713 MRR. Removing context's local matching and matching composition (``-CtxDec'') degraded R@10 and MRR to 0.117 and 0.0620. Ensembling the above two kinds of ESIM methods (``Ensemble2'') achieved 0.134 R@10 and 0.0770 MRR.

For Ubuntu subtask 4, we observed similar trend with subtask 1. ESIM achieved 0.887 R@10 and 0.6434 MRR, ``-CtxDec'' degraded performance to 0.877 R@10 and 0.6277 MRR, and ``-CtxDec \& -Rev'' further degraded to 0.875 R@10 and 0.6212 MRR. Ensembling the above three models (``Ensemble'') achieved 0.909 R@10 and 0.6771 MRR. 

For Ubuntu subtask 5, the dataset is the same as subtask 1 except for using the external knowledge of Linux manual pages. Adding pre-trained word embeddings derived from Linux manual pages (``+W2V'') resulted in 0.858 R@10 and 0.6394 MRR, comparable with ESIM without exploring the external knowledge. Ensembling the ensemble model for subtask 1 (0.887 R@10 and 0.6790 MRR) and the "+W2V" model brought further gain, reaching 0.890 R@10 and 0.6817 MRR.

 Table~\ref{tab:result:advising} showed the ablation analysis on the development set for the Advising dataset. We used ESIM without context's local matching and matching composition for computational efficiency. We observed similar trends as on the Ubuntu data set. ``-CtxDec \& -Rev'' degraded R@10 and MRR over ``-CtxDec'', yet the ensemble of the two models always produced significant gains over individual models.

\subsection{Comparison with Previous Work}

The results on two public benchmarks were summarized in Table~\ref{tab:result:public}. The first group of models includes sentence-encoding based methods. They use hand-craft features or neural network features to encode both context and response, then a cosine classifier or MLP classifier was applied to decide the relationship between the two sequences. Previous work used TF-IDF, RNN \cite{DBLP:conf/sigdial/LowePSP15} and CNN, LSTM, BiLSTM \cite{DBLP:journals/corr/KadlecSK15} to encode the context and the response. 

The second group of models consists of sequence-based matching models, which usually use the attention mechanism, including MV-LSTM~\cite{DBLP:conf/aaai/WanLGXPC16}, Matching-LSTM~\cite{DBLP:conf/naacl/WangJ16}, Attentive-LSTM~\cite{DBLP:journals/corr/TanXZ15}, and Multi-Channels~\cite{DBLP:conf/acl/WuWXZL17}. These models compared the token-level relationship between the context and the response, rather than comparing the two dense vectors directly as in sentence-encoding based methods. These kinds of models achieved significantly better performance than the first group of models. 

The third group of models includes more complicated hierarchy-based models, which usually models the token-level and utterance-level information explicitly. Multi-View~\cite{DBLP:conf/emnlp/ZhouDWZYTLY16} model utilized utterance relationships from the word sequence view and utterance sequence view. DL2R model~\cite{DBLP:conf/sigir/YanSW16} employed neural networks to reformulate the last utterance with other utterances in the context. SMN model~\cite{DBLP:conf/acl/WuWXZL17} used CNN and attention to match a response with each utterance in the context. DUA~\cite{DBLP:conf/coling/ZhangLZZL18} and DAM~\cite{DBLP:conf/acl/WuLCZDYZL18} applied a similar framework as SMN~\cite{DBLP:conf/acl/WuWXZL17}, where one improved with gated self attention and the other improved with the Transformer~\cite{DBLP:conf/nips/VaswaniSPUJGKP17}. 

Although the previous hierarchy-based work claimed that they achieved the state-of-the-art performance by using the hierarchical structure of multi-turn context, our ESIM sequential matching model outperformed all previous models, including hierarchy-based models. On the Lowe's Ubuntu dataset, the ESIM model brought significant gains on performance over the previous best results from the DAM model, up to 79.6\% (from 76.7\%) R@1, 89.4\% (from 87.4\%) R@2 and 97.5\% (from 96.9\%) R@5. For the E-commerce dataset, the ESIM model also accomplished substantial improvement over the previous state of the art by the DUA model, up to 57.0\% (from 50.1\%) R@1, 76.7\% (from 70.0\%) R@2 and 94.8\% (from 92.1\%) R@5. These results demonstrated the effectiveness of the ESIM model, a sequential matching method, for multi-turn response selection.

\section{Conclusion}
Previous state-of-the-art multi-turn response selection models used hierarchy-based (utterance-level and token-level) neural networks  to explicitly model the interactions among the different turns' utterances for context modeling. In this paper, we demonstrated that a sequential matching model based only on chain sequence can outperform all previous models, including hierarchy-based methods, suggesting that the potentials of such sequential matching approaches have not been fully exploited in the past. Specially, the proposed model achieved top one results on both datasets under the noetic end-to-end response selection challenge in DSTC7, and yielded new state-of-the-art performances on two large-scale public multi-turn response selection benchmarks. Future work on multi-turn response selection includes exploring the efficacy of external knowledge~\cite{DBLP:conf/acl/InkpenZLCW18}, such as knowledge graph and user profile. 

\bibliographystyle{aaai}
\bibliography{refs}

\end{document}